\documentclass{article}
\usepackage{spconf,amsmath,graphicx}

\usepackage{epsfig, amstext,amsfonts,amssymb,subfigure,subeqnarray,nccfoots,color,multirow,bm, indentfirst, enumitem}

\usepackage{booktabs}
\usepackage{graphicx}
\usepackage{subfigure}
\usepackage[toc,page]{appendix}
\usepackage{float}
\usepackage{threeparttable}

\DeclareMathOperator*{\argmin}{argmin}
\newcommand{\norm}[1]{\left\lVert#1\right\rVert}
\title{How Effectively Can Indoor Wireless Positioning Relieve Visual Tracking Pains: A Cramer-Rao Bound Viewpoint}
\name{Panwen Hu$^{\star}$ \qquad Zizheng Yan$^{\star}$ \qquad Rui Huang$^{\star }$ \qquad Feng Yin$^{\star \dagger *}$
\thanks{*Feng Yin is the correspondence author.}}
\address{$^{\star}$School of Science and Engineering, The Chinese University of Hong Kong, Shenzhen, P.R. China \\$^{\dagger}$Shenzhen Research Institute of Big Data (SRIBD), Shenzhen, Guangdong, 518172, P.R. China}

\begin{document}
\ninept
\maketitle
\begin{abstract}
Visual tracking is fragile in some difficult scenarios, for instance, appearance ambiguity and variation, occlusion can easily degrade most of visual trackers to some extent. In this paper, visual tracking is empowered with wireless positioning to achieve high accuracy while maintaining robustness. Fundamentally different from the previous works, this study does not involve any specific wireless positioning algorithms. Instead, we use the confidence region derived from the wireless positioning Cram\'{e}r-Rao bound (CRB) as the search region of visual trackers. The proposed framework is low-cost and very simple to implement, yet readily leads to enhanced and robustified visual tracking performance in difficult scenarios as corroborated by our experimental results. Most importantly, it is utmost valuable for the practioners to pre-evaluate how effectively can the wireless resources available at hand alleviate the visual tracking pains.
\end{abstract}
\section{Introduction}
\label{sec:intro}

Visual tracking is one of the core research fields in computer vision, it has wide applications in video data analytics, automatic driving, etc. Although visual tracking research has made much remarkable progress in the past decades, there are still some difficult scenarios in which it may work improperly. Some difficulties that visual tracking nowadays is facing include, among others, 1) intra-class variations in targets' appearances, such as deformation, scaling, poses, etc. 2) inter-class variations between targets, such as ambiguities in association, identity switching, etc. 3) variations in the scenes, such as background clutters, differences in multi-camera views, occlusion, etc.

As pointed out in \cite{Alahi2015}, wireless data and visual data are complementary and should be fused for enhanced tracking performance. In the ideal cases where the above mentioned hurdles do not occur, visual tracking is in general more accurate and informative than wireless positioning. However, when any of these challenges are present, the performance of visual tracking will degenerate to some extent, while wireless positioning is much less influenced. The combination of the two can guarantee both the positioning accuracy and robustness in most scenarios.

%Although the existing works have shown the effectiveness of incorporating wireless positioning in various scenarios, their application scenario is narrow since their fusion strategy can only be applied on one or two positioning infrastructures and trackers.
In this work, we mainly focus on \textbf{long-term multi-camera single-object tracking}, but the idea can be extended to the multi-object case straightforwardly. Different from the existing works, we propose a general framework utilizing wireless positioning to empower a broad range of visual trackers. \textbf{More concretely, our contributions are as follows. First, we propose to empower a visual tracker by replacing its empirical search region with a more reliable confidence region constructed by the wireless positioning Cram\'{e}r-Rao bound (CRB), which is simple to compute.
%Second, the proposed framework is able to perform multi-camera tracking without the detection or projection step, which makes it more lightweight than its competitors.
Second, the proposed framework does not specify any sophisticated wireless positioning algorithms, and more importantly, it can be used to evaluate the best achievable tracking performance in terms of the given type and number of wireless devices as well as the quality of the wireless measurements. This is utmost valuable for practioners to pre-evaluate if a desired visual tracking system can achieve a designated goal given all available wireless resources at hand, or on the other hand, to choose proper wireless positioning solutions based on the desired tracking performance. Lastly, we publish a benchmark dataset for the readers to evaluate different algorithms.}

The rest of this paper is organized as follows. Section~\ref{sec:relatedwork} reviews related works on visual tracking and wireless positioning. Section~\ref{sec:Framework} introduces the proposed confidence region based visual tracking framework. Section~\ref{sec:experiments} shows some experimental results. Finally, Section~\ref{sec:Conclusion} concludes the paper.

\section{Related Work}
\label{sec:relatedwork}

\subsection{Visual Tracking}
Visual trackers can be broadly categorized into two classes \cite{Smeulders2013}: generative trackers \cite{Oron2012,Mei2015} and discriminative tracker \cite{Danelljan2016,Song2017,Fan2017a,Zhang2017}. Generative trackers formulate the tracking process as searching for image regions whose extracted features are most similar to those of the target, whereas discriminative trackers aim at separating the target from background by training classifiers with specific data. No matter for generative or discriminative trackers, robust sampling strategies, i.e., motion models in \cite{Wang2015}, can alleviate some of the visual tracking pains mentioned before and are becoming important. Motion models employed by canonical trackers are diverse, for instance,  Struck \cite{Hare2011} employes the radius sampling strategy around the predicted location of the previous frame; ECO \cite{Danelljan2016}, CREST \cite{Song2017}, LMCF \cite{Wang2017} use patches, centered at the target position of previous frame and \textbf{k}-times the target size, as the search region; LOT \cite{Oron2012}, L1T \cite{Mei2015}, RaF \cite{Zhang2017}, SANet \cite{Fan2017a} draw samples with particle filters  \cite{Arulampalam2002}. However, all of these sampling methods depend heavily on the tracking results of previous frames, even when the previous results are inaccurate or even erroneous.

\subsection{Wireless Positioning}
Wireless positioning has attracted a lot of attention in the last two decades. A plethora of deterministic and stochastic indoor positioning approaches based on different position-related measurements have been proposed. Table~1 compares a few most recent and representative ones from various aspects. More related works can be found in their references.

\begin{table*}[htbp]
  \centering
  \caption{Representative Positioning Approaches with Different Measurements.}
    \begin{tabular}{|l|l|l|l|l|l|}
    \hline
    \textbf{measurement} & \textbf{accuracy (basic setup)} & \textbf{device complexity} & \textbf{deploy cost} &\textbf{ NLOS  impact} & \textbf{supported infrastruce }\\
    %\hline
    %\textcolor[rgb]{ .267,  .447,  .769}{sattelite} & $>$10 meter or NA & medium & severe & GPS,Beidou\\
    %\hline
    %\textcolor[rgb]{ .267,  .447,  .769}{proximity\cite{Yin2016}} & ~5 meter to 2 meter  & low   & moderate & BLE, WiFI, LTE/5G\\
    \hline
    \textcolor[rgb]{ .267,  .447,  .769}{RSS\cite{Yin2017}} & ~5 meter to 2 meter  & low & low  & moderate & BLE, WiFI, LTE/5G\\
    %\hline
    %\textcolor[rgb]{ .267,  .447,  .769}{AoA\cite{Garcia2017}} & 2 meter to submeter & medium & moderate & BLE, WiFI, LTE/5G\\
    \hline
    \textcolor[rgb]{ .267,  .447,  .769}{TOA\cite{Yin2013}} & 2 meter to submeter & medium & high & moderate & WiFi,Cellular\\
    \hline
    \textcolor[rgb]{ .267,  .447,  .769}{UWB\cite{Zwirello2012}} & submeter  to centimeter & medium & high & moderate & UWB\\
    \hline
    \textcolor[rgb]{ .267,  .447,  .769}{CSI\cite{Wang2017a} } & submeter to decimeter & high & high & moderate & WiFi, 5G\\
    \hline
    \textcolor[rgb]{ .267,  .447,  .769}{Visible Light\cite{Zhu2017}} & decimeter to centimeter & high & high & very severe & LiFi\\
    \hline
    \end{tabular}%
  \label{tab:PositioningAlgorithms}%
\end{table*}%

\subsection{Wireless Positioning Assisted Visual Tracking}
Surprisingly, there are very few existing works on combining wireless positioning with visual tracking. In \cite{Alahi2015}, visual and wireless data are jointly processed in a single-camera multi-object tracking scenario to alleviate the association problem. While our proposed framework is suitable for both single-camera and multi-camera scenarios. In \cite{Zhu2011}, the authors use two directional antennae to estimate the position of a target person first in world coordinates, and then plug this position information into a particle filter to further improve the tracker's performance. In \cite{Pham2017}, a novel WiFi and visual data fusion scheme is proposed to get the target person's position. Specifically, a Kalman filter is used for tracking, while a data classifier with kernel descriptors is used for person Re-IDentification (Re-ID). Comparing these approaches directly is somewhat meaningless, because they were designed specially for different scenarios. \textbf{Our work is fundamentally different from the above ones in the sense that we do not aim at any particular tracking algorithms, instead, we place more emphasis on searching for a general framework that both enables enhanced tracking performance using wireless positioning and, more importantly, provides a principled guidance for designing the overall tracking system.}

\section{Confidence Region Based Visual Tracking}
\label{sec:Framework}
\subsection{A Full Picture}
\begin{figure}[htbp]
  \centering
  \includegraphics[width=7.5cm]{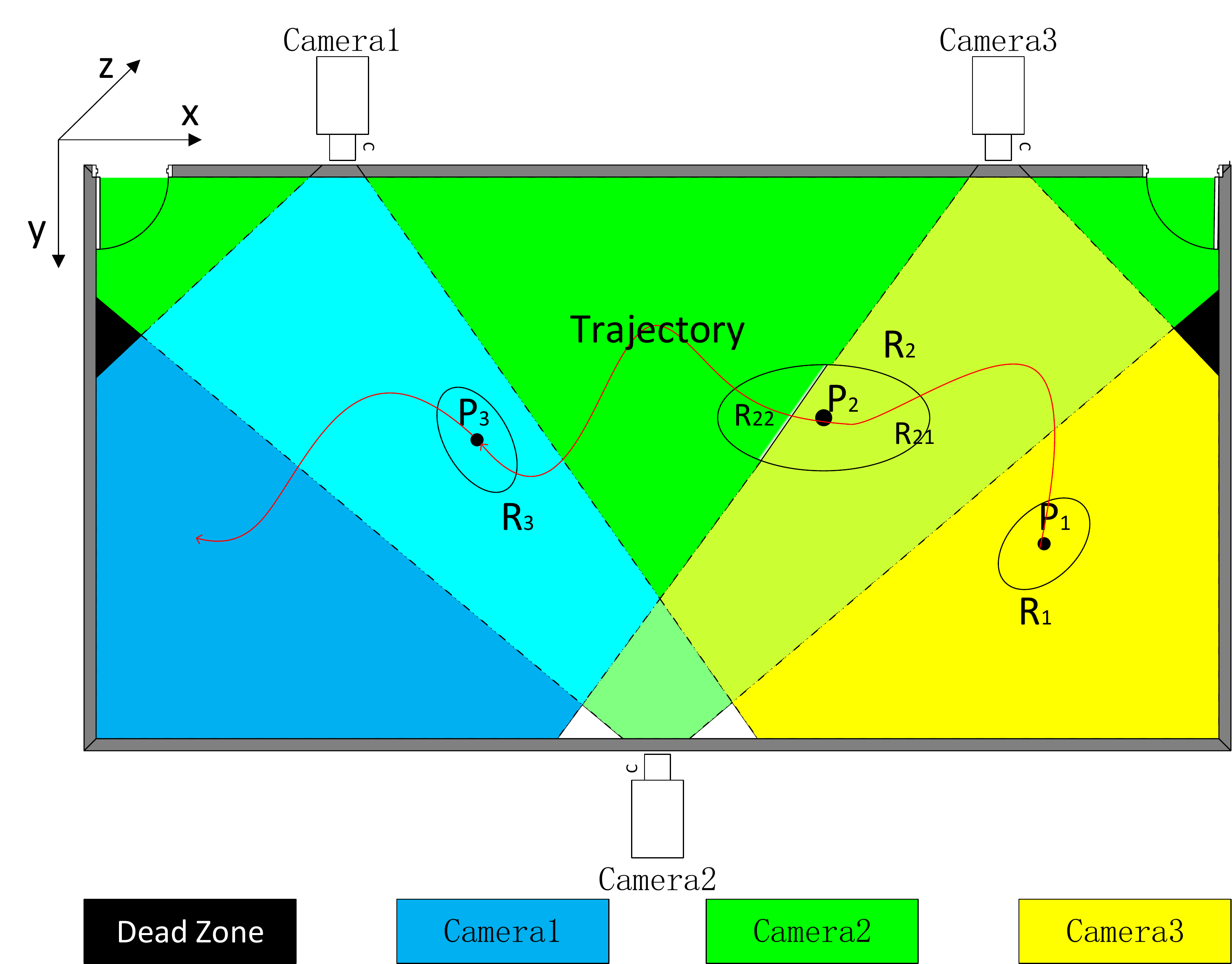}
  \caption{Confidence region based  multi-camera single-object tracking. View of interest (VOI) is enclosed by dash lines and the red curve from $P_1$ to $P_3$ is the trajectory. $R_1$, $R_2$, $R_3$ are the corresponding confidence regions determined by the wireless positioning (better viewed in color).
%When the target moves to $P_2$, leaving camera 3 and entering camera 2, search regions will be formulated in both camera 2 and camera 3.
}
  \label{fig:scenario}
\end{figure}

In this section, we present the confidence region based visual tracking framework in details. We first give a full picture of applying the confidence region based visual tracking to long-term multi-camera single-object tracking in Fig.~\ref{fig:scenario}.
%In most traditional visual trackers, a rectangle region centered at the target is used as the search region of next frame.
In the proposed framework, the confidence region derived from the wireless positioning Cram\'{e}r-Rao bound (CRB) is used as the search region of a visual tracker. As a consequence, well enhanced tracking performance can be obtained even for difficult scenarios mentioned in Section~\ref{sec:intro}. For instance, in the case of data association, traditional trackers may have traced a wrong target in the current frame and will use the wrong search region for all subsequent frames. In contrast, the search region in our framework is obtained from wireless positioning, which is self-healing and more reliable. In the next subsection, we will show how to compute the confidence region from wireless positioning.

\subsection{Confidence Region Obtained from Wireless Positioning}
\label{subsec:confidenceReg}
Before the confidence region is derived, we make the following assumptions: 1) A number of $N$ wireless sensors are deployed in the area of interest with precisely known geographical position $\boldsymbol{p}_{i} \triangleq [x_i, y_i, z_i]^T, i=1,2,...,N$. These sensors are often called anchors. 2) The wireless sensors can receive position related signals regularly from a wireless transmitter worn by a target person. The measurement collected from the $i$th sensor at a sampling time instance can be expressed in general form as follows:
	\begin{equation}
	r_{i} = h_{i}\left( \boldsymbol{p}, \boldsymbol{p}_{i}, \boldsymbol{\theta}_m \right) + v_i,
	\end{equation}
where $r_{i}$ is the received signal at the $i$th anchor; $h_{i}(\cdot)$ is often a nonlinear function in terms of $\boldsymbol{p} \triangleq [x, y, z]^T$, i.e., the position of the wireless transmitter (or equivalently the target person) to be located; the anchor positions $\boldsymbol{p}_{i}$, $i=1,2,...,N$ and the measurement model parameters $\boldsymbol{\theta}_m$ are deemed as known due to the calibration; $v_i$ is the corresponding noise term in the $i$-th measurement $r_i$. The signal model can be written in vector form as follows:
\begin{eqnarray}
\label{eqn:sigmal-model-vector-form}
\boldsymbol{r} =  \boldsymbol{h}(\boldsymbol{p}) + \boldsymbol{v},
\end{eqnarray}
where $\boldsymbol{r} = [ r_{1},..., r_{N} ]^{T}$, $\boldsymbol{h}(\boldsymbol{p}) = [ h_{1}\left( \boldsymbol{p} \right),..., h_{N}\left( \boldsymbol{p}\right) ]^{T}$ and $\boldsymbol{v} = [ v_{1},...,v_{N}]^{T}$ that are all of dimension $N \times 1$. 3) The measurement noise $v_i$, $i=1,2,...,N$ are independently and identically distributed according to $p(v)$, which is assumed to be known. In case $p(v)$ is unknown, a Gaussian distribution will be used for approximation. 4) The $z$ dimension of the confidence region is set to $0$ in world coordinates.

\textbf{Theorem 1}: The Fisher Information Matrix (FIM) of any unbiased static position estimator can be derived, in light of \cite{Yin2013}, as
\begin{equation}
\label{eqn:FIM-expression-afterchainrule}
\boldsymbol{\mathcal{F}}(\boldsymbol{p}) = \mathcal{I}_{v} \cdot \boldsymbol{\mathcal{H}}(\boldsymbol{p})  \boldsymbol{\mathcal{H}}^{T}(\boldsymbol{p}),
\end{equation}
where $\boldsymbol{\mathcal{H}}(\boldsymbol{p}) = \nabla_{\boldsymbol{p}} \boldsymbol{h}(\boldsymbol{p})$
%
%\begin{equation}
%\boldsymbol{\mathcal{H}}(\boldsymbol{p}) = \nabla_{\boldsymbol{p}} \boldsymbol{h}(\boldsymbol{p})
%\end{equation}
and $\mathcal{I}_{v} = \mathbb{E}_{p(v)} \left\lbrace \frac{\left[\nabla_{v} p(v)\right]^2}{p^{2}(v)} \right\rbrace$.
%$\mathcal{I}_{v}$ is calculated by
%\begin{equation}
%\label{eqn:IA-scaling-factor}
%\mathcal{I}_{v} = \mathbb{E}_{p_{V}(v)} \left\lbrace \frac{\left[\nabla_{v} p_{V}(v)\right]^2}{p^{2}_{V}(v)} \right\rbrace  = \int \frac{\left[\nabla_{v} p_{V}(v)\right]^2}{p^{2}_{V}(v)} p_{V}(v) \rm{d}v
%\end{equation}

Often, the scaling factor $\mathcal{I}_{v}$ has to be evaluated numerically via Monte Carlo integration, except for the Gaussian distributed noise, for which $\mathcal{I}_{v}$ can be derived in closed form and is equal to the inverse of the noise variance, $\sigma_{v}^2$.

When the regularity conditions \cite{Gustafsson2010} are all fulfilled, the covariance matrix of any unbiased estimator $\hat{\boldsymbol{p}}$ satisfies
\begin{equation}
{\rm{Cov}}(\hat{\boldsymbol{p}}) = \mathbb{E}_{p(\boldsymbol{r};\boldsymbol{p})}\left\lbrace (\hat{\boldsymbol{p}}-\boldsymbol{p})(\hat{\boldsymbol{p}}-\boldsymbol{p})^{T}\right\rbrace \succeq \boldsymbol{\mathcal{F}}^{-1}(\bm{p}),
\end{equation}
where $\boldsymbol{\mathcal{F}}(\boldsymbol{p})$ denotes the Fisher's information matrix (FIM) given in Theorem 1.

%Remark I: The best case unbiased estimate provides $E(\hat{x}) = x_{*}$ and $var(\hat{x}) = \left[ \boldsymbol{\mathcal{F}}^{-1}(\bm{\theta}) \right]_{11}$ for the $x$-direction and $E(\hat{y}) = y_{*}$ and $var(\hat{y}) = \left[ \boldsymbol{\mathcal{F}}^{-1}(\bm{\theta}) \right]_{22}$ for the $y$-direction.

\textbf{Definition 1}: For mathematical tractability, we define a $(1-\alpha)\times 100\%$ percentage elliptical confidence region (a function in terms of the position $\boldsymbol{p}$) by
\begin{equation}
\chi_{2}^{2}(\alpha) \triangleq \left( \boldsymbol{p} - \boldsymbol{p}_{b} \right)^{T} \boldsymbol{\mathcal{F}}(\boldsymbol{p})  \left( \boldsymbol{p} - \boldsymbol{p}_{b} \right),
\label{eqn:EllipticalConfidenceRegion}
\end{equation}
where $\boldsymbol{p}_{b}$ represents the boundary of a given confidence region.

\textbf{Remark}: The $(1-\alpha)\times 100\%$ percentage confidence region defined above corresponds to an unbiased, multivariate Gaussian distributed position estimator that achieves the CRB. In practice, such ideal position estimator may not exist, but some estimators may perform fairly close to it, especially when the number of measurements is large; for instance, the maximum likelihood estimator is asymptotically efficient \cite{Kay1993}.

In the last step, perspective projection is adopted to project the confidence region in Eq.(\ref{eqn:EllipticalConfidenceRegion}) onto a 2D image plane. To be precise, the projection from world coordinates to pixel coordinates, in light of \cite{Zhang2000}, is
\begin{equation}
\label{eqn:transformation}
  s
  \begin{bmatrix}
    x_{p} \\
    y_{p} \\
    1
  \end{bmatrix}
  = \boldsymbol{K}
  \begin{bmatrix}
    \boldsymbol{R} & \boldsymbol{T}
  \end{bmatrix}
  \begin{bmatrix}
    x_{w} \\
    y_{w} \\
    z_{w} \\
    1
  \end{bmatrix},
\end{equation}
%with
%\[
%    \boldsymbol{K}=\begin{bmatrix}
%                     f_{x} & 0 & u_{0} \\
%                     0 & f_{y} & v_{0} \\
%                     0& 0& 1
%                   \end{bmatrix}
%\]
%
where $(x_{p},y_{p})$ is a point in pixel coordinates, $(x_{w},y_{w},z_{w})$ is a point in world coordinates, $\boldsymbol{R} \in \mathbb{R}^{3 \times 3}$ is an extrinsic parameter matrix, $\boldsymbol{T} \in \mathbb{R}^{3 \times 1}$ is a translation vector, which relates to the projection from world coordinates to camera coordinates, $\boldsymbol{K} \in \mathbb{R}^{3 \times 3}$ is called camera's intrinsic parameter matrix, which is used to transform 3D points in camera coordinates to pixels coordinates, $s$ is an scale factor depending on $\boldsymbol{R}$ and $\boldsymbol{T}$. The parameters $s$, $\boldsymbol{R}$, $\boldsymbol{K}$ and $\boldsymbol{T}$ will be calibrated as soon as the visual tracking system is installed. Now the motion model of visual trackers can be estimated by the confidence region on the 2D image plane.

\textbf{Example}: Consider a bluetooth-low-energy (BLE) network with $N$ beacons and a visual tracking system co-locate in the deployed area. Visual data and wireless data are both used for target tracking. The wireless BLE measurement noise terms are assumed to be Gaussian i.i.d. with zero mean and variance equal to $\sigma_{v}^2$. We adopt the log-distance path-loss model
\begin{equation}
h_{i}\left( \boldsymbol{p}, \boldsymbol{p}_{i}, \boldsymbol{\theta}_m \right) = A + 10 B
\log_{10}\left( || \boldsymbol{p} - \boldsymbol{p}_{i}||\right), \, i=1,2,...,N,
\end{equation}
where $\boldsymbol{\theta}_m \triangleq [A, B]^{T}$ is a set of calibrated propagation model parameters which may remain different across different anchors.

For this scenario, we can easily derive $\boldsymbol{\mathcal{H}}(\boldsymbol{p})$ and $\mathcal{I}_{v}$ in closed form as follows:
\begin{equation}
\begin{split}
\boldsymbol{\mathcal{H}}(\boldsymbol{p})  = \nabla_{\boldsymbol{p}} \boldsymbol{h}(\boldsymbol{p}) = \frac{10}{\ln 10} \begin{bmatrix}
\frac{B_1(\boldsymbol{p}-\boldsymbol{p}_1)}{d_1^2} ...  \frac{B_N(\boldsymbol{p}-\boldsymbol{p}_N)}{d_N^2}
\end{bmatrix}
\end{split},
\end{equation}
where $d_i^2 = (\boldsymbol{p}-\boldsymbol{p}_i)^T (\boldsymbol{p}-\boldsymbol{p}_i)$ and $\mathcal{I}_{v} = \sigma_{v}^{-2}$ for zero mean Gaussian distributed noise with variance $\sigma_{v}^2$. With these results, a confidence region can be constructed easily with Eq.(\ref{eqn:FIM-expression-afterchainrule}) and Eq.(\ref{eqn:EllipticalConfidenceRegion}). Lastly, the object's motion in pixel coordinate can be estimated by projecting the confidence region in world coordinate to the 2D image plane with Eq.(\ref{eqn:transformation}). It is noteworthy that the FIM should have been evaluated with the true position, $\boldsymbol{p}$, but due to the lack of the ground truth in practice, it is often replaced with a position estimate, for instance, the maximum likelihood estimate. Due to the Gaussian noise, the ML estimate is equivalent to the least-squares estimate \cite{Kay1993},
\begin{equation}
\hat{\boldsymbol{p}}_{LS} =  \argmin_{\boldsymbol{p}} \sum_{i=1}^{N} \left( r_i-A_i-10B_i \log_{10}(\norm{\boldsymbol{p} - \boldsymbol{p_i}}) \right)^2.
\end{equation}
A nonlinear programming solver, e.g., Newton's method and conjugate gradient method \cite{Bertsekas16}, can be used to solve this problem.

\section{Experimental Results}
\label{sec:experiments}
\subsection{Data and General Setup}
\begin{figure*}
	\centering
	\includegraphics[width = 14cm,height=6cm]{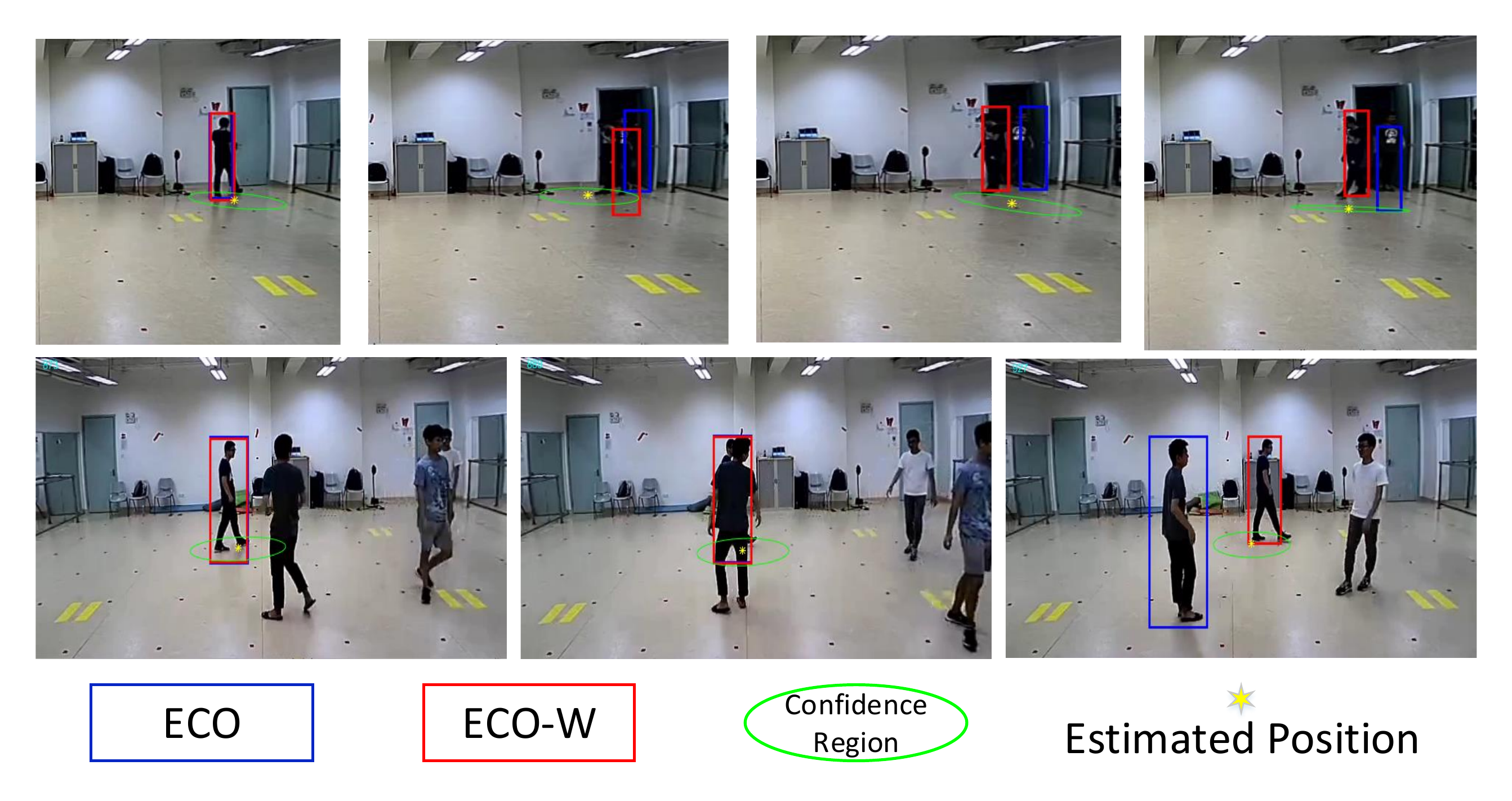}
	\caption{The first row illustrates that our new model can resume tracking from a failure caused by the strong background clutter. The second row illustrates the occlusion by a similar object, the proposed confidence region based model can track the right person after reappearing. \textbf{In both cases, the original ECO tracker failed.} Note that in our experiments, we use \textcolor{red}{95\%} confidence region.(better viewed in color)}
	\label{fig:Comparison}
\end{figure*}

To the best of our knowledge, there is no public dataset available so far, which contains wireless data, camera parameters, visual data along with annotations of the object collected in difficult scenarios. In order to evaluate the proposed confidence region-based visual tracking framework, we collected 3 video clips with in total 21600 frames, shot by 3 carefully calibrated cameras deployed as in Fig.~\ref{fig:scenario}. This real dataset contains various different visual tracking challenges, including occlusion, visual ambiguity, appearance variations, etc. Figure.\ref{fig:Comparison} shows some examples of the dataset. Complete dataset can be found on our website.\footnote{https://github.com/w-tracking/w-tracking-dataset}

In order to study how the goodness of wireless positioning will impact the visual tracking performance, we consider a BLE network with $N=32$ beacons and simulate Receive Signal Strength (RSS) measurements according to the model given in Section~\ref{sec:Framework} subject to Gaussian noise with zero mean and different noise variances $\sigma_{v}^2$. %Since positioning accuracy is dependent on both number and variance of wireless nodes, the combination of different number and variance of wireless nodes will be very redundant.
We use the Mean-Square-Error (MSE) as a measure of wireless positioning accuracy. Table~\ref{tab:locationError} shows the positioning MSE versus the noise standard deviation $\sigma_v$. A larger positioning MSE value also means a bigger confidence region derived from the CRB, which is simple to compute. In the sequel, we mainly compare the confidence region based ECO model (named ECO-W for short) with the state-of-the-art ECO model \cite{Danelljan2016} without wireless positioning.

\begin{table}[H]
	\centering
	\caption{Positioning MSE versus RSS noise standard deviation.}
	\begin{tabular}{|c|c|c|c|c|c|c|}
		\hline
		$\sigma_v$, std (dBm)    & 3     & 5     & 7     & 9     & 11   \\
		\hline
		Accuracy(cm)  & 36 & 63 & 92 & 121 & 152  \\
		\hline
	\end{tabular}%
	\label{tab:locationError}%
\end{table}

\subsection{Results}
\label{sec:Experimental-results}
%We have integrated our fusion framework into visual tracking system (called fusion system in the following) to obtain better performance when conduct single-object long-term tracking under the multi-camera scene.
%In order to explore the impact of wireless positioning accuracies on tracking performances, we simulated the RSS using algorithm in \cite{Yin2013} with different accuracies. Since positioning accuracy is dependent on both number and variance of wireless nodes, the combination of different number and variance of wireless nodes will be very redundant. So we just use MSE (Mean Square Error) of wireless positioning evaluated in our trajectory as the accuracy metric of wireless positioning, as displayed in Table \ref{tab:locationError}.
%and the corresponding positioning errors to \textbf{raw annotation}\footnote{we could not obtain exactly 3D location coordinates for observed object with simple devices like cameras, beacons, ect.}

The following experiments are conducted for long-term multi-camera single-object tracking. First, we measure robustness in terms of recall rate (the proportion, in percentage, of samples that contain target individuals out of all samples) obtained by the ECO model versus the proposed ECO-W model. The results shown in Table~\ref{tab:Recallrate} confirm that the ECO-W model has achieved higher recall rates than that of the ECO model, despite of the high noise level in the RSS measurements.
\begin{table}[h]
	\centering
	\caption{Recall rate of ECO model versus ECO-W model.}
	\begin{tabular}{|c|c|c|c|c|c|}
		\hline
		$\sigma_v$ (dBm) & 3     & 5     & 7     & 9     & 11   \\
		\hline
		ECO-W & 99.9 \% & 99.8  \% & 98.6  \% & 95.7  \% & 92.3  \% \\
		\hline
       ECO & \multicolumn{5}{|c|}{91.1  \%} \\
        \hline
	\end{tabular}%
\label{tab:Recallrate}%
\end{table}

\begin{figure}[htb]
	\centering
	\subfigure[Success plot]{
		\label{fig:ECOHCsubfig:a}
		\begin{minipage}{7cm}
			\centering
			\includegraphics[scale = 0.5]{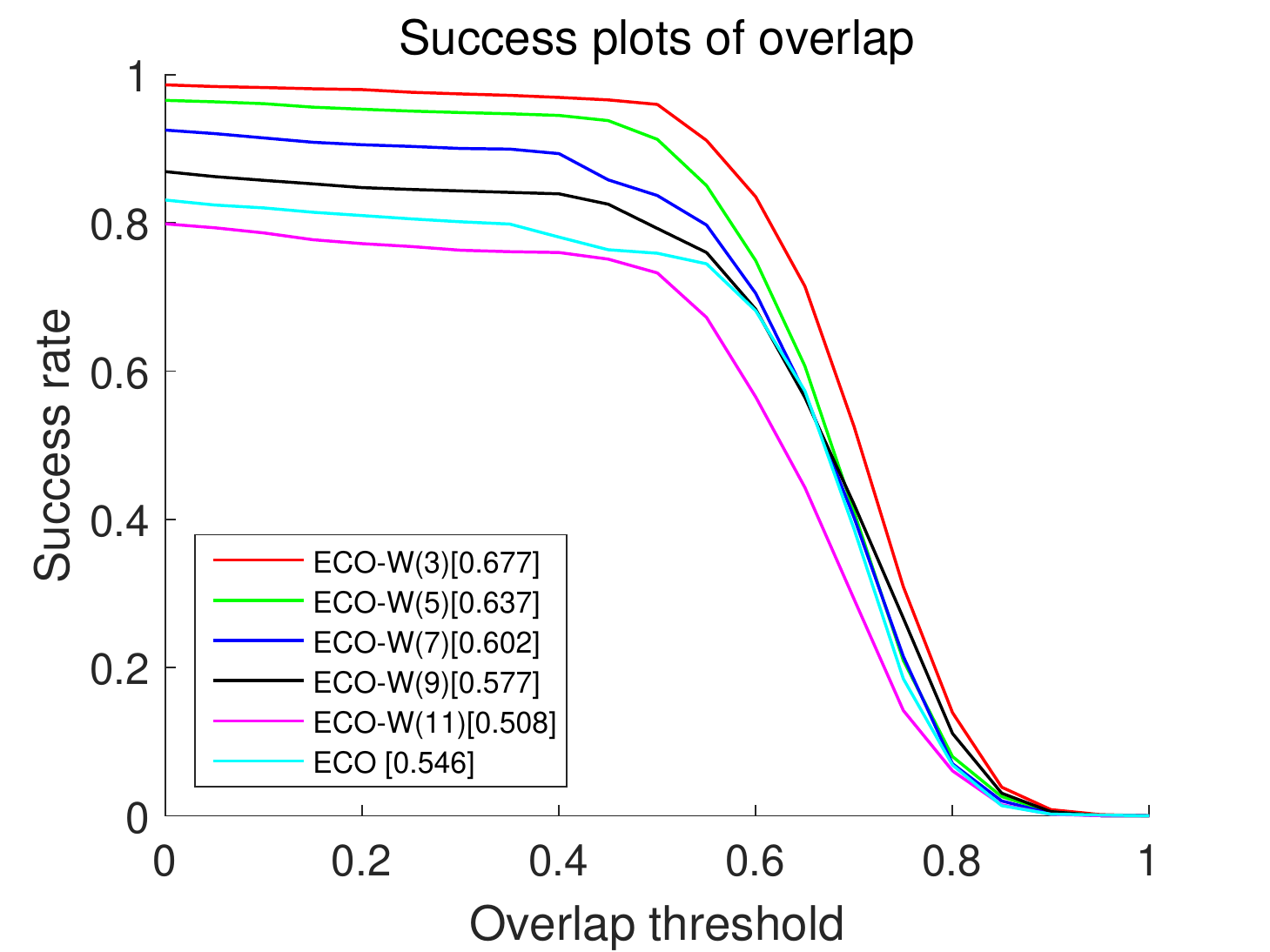}
		\end{minipage}
	}

	\caption{Overlap success rate of the ECO model versus ECO-W model. The number in a round bracket $(\cdot)$ is the measurement noise standard deviation. The number in a square brackets $[\cdot]$ is the corresponding AUC score.}
\label{fig:ECO_HCplots}	
\end{figure}

In the second experiment, we utilize Overlap Success Rate (OSR) as was used in \cite{Wu2013a} to evaluate the tracking performance of ECO-W model versus the ECO model. Here, OSR is the ratio of frames whose overlap scores, computed from the unions and intersections of tracked bounding boxes and ground-truth boxes, are larger than a given threshold. Then, Area Under Curve (AUC) is computed to summarize and rank the overall performance of the two visual trackers. Figure \ref{fig:ECO_HCplots} shows the OSR of the ECO-W model versus the ECO model for different positioning MSE values.

\textbf{The key message from the two experiments is that the ECO-W model outperforms the ECO model when the positioning MSE is smaller than $150 cm$ (may vary for different camera settings) in terms of the AUC score.} Otherwise, wireless positioning may influence a visual tracking system negatively if the MSE is too large. Avoiding this requires one either use more sophisticated wireless devices with great maintenance effort or use larger-scale network of cheap and crude wireless devices. Compared with the trackers using pure visual information, our framework takes advantages of wireless positioning to correct the bounding box of the target when two or more similar objects separated after being wrongly traced, see Fig.~\ref{fig:Comparison} for some examples. Wireless positioning data provides a more accurately confined and more reliable search region, which is the essential condition for visual trackers to avoid model drifting and identity switching, and achieve improved tracking performance in long-term tracking.

Another important observation is that the proposed framework does not depend on any specific wireless network setup (including network type, measurement type, device cost, etc.). It helps practitioners to pre-compute the best achievable MSE and will immediately return an evaluation on the feasibility given the available type and number of wireless devices, the measurement quality, etc. On the other hand, although the best achievable MSE is optimistic, it is still informative and tells the practitioners how well a specific wireless positioning solution should be designed e.g., how many devices should be purchased and how many measurements (depending on their qualities) should be collected so that the final wireless assisted visual tracking system can meet all design specifications.

\section{Conclusion}
\label{sec:Conclusion}
In this work, we proposed a general statistical framework to fuse visual data with wireless data, which demonstrated itself to be more accurate and more robust for target tracking in some difficult scenarios. The proposed framework provided not only a novel, practical, low-cost, lightweight wireless assisted visual tracker but also a principled guidance for selecting and evaluating the effectiveness of a wireless network setup, including the network type, wireless devices and measurement quality, etc. Experimental results demonstrated that some visual tracking pains, such as long-term tracking, feature model drifting, recovery, and so on could be alleviated effectively when the wireless positioning accuracy is good enough.

%\vfill
%\pagebreak

\bibliographystyle{IEEEbib}
\bibliography{ref}

\end{document}